%% file: main.tex
\documentclass[letterpaper, 10 pt, journal, twoside]{IEEEtran}
\usepackage{graphicx}
\usepackage{enumitem}
\usepackage{cite}
\usepackage{amsmath,amsfonts, amssymb}
\usepackage{algorithmic}
\usepackage{algorithm}
\usepackage{array}
\usepackage[caption=false,font=normalsize,labelfont=sf,textfont=sf]{subfig}
\usepackage{textcomp}
\usepackage{stfloats}
\usepackage{url}
\usepackage{verbatim}
\usepackage{booktabs}
\usepackage{graphicx}
\usepackage{cite}
\usepackage{xspace}
\usepackage{kotex}
\usepackage{xcolor}
\usepackage{multirow}
\usepackage{comment}
\usepackage{hyperref}
\usepackage{enumitem}
\usepackage{cite}
\hypersetup{
    colorlinks=true,       
    linkcolor=magenta,     
    urlcolor=magenta       
}

\newcommand{\kush}[1]{\textcolor{black}{#1}}
\newcommand{\kushtimusprime}[1]{\textcolor{black}{#1}}

\ifCLASSINFOpdf
\else
\fi
\begin{document}
%
\title{STITCH 2.0: Extending Augmented Suturing with EKF Needle Estimation and Thread Management}
%
%
%

\author{Kush Hari$^{1}$, Ziyang Chen$^{1}$, Hansoul Kim$^{1}$, and Ken Goldberg$^{1}$%
\\ 
\href{https://stitch-2.github.io/}{https://stitch-2.github.io}
\vspace{-1.1em}
\thanks{Manuscript received: July, 10, 2025; Accepted October, 3, 2025.}
\thanks{This paper was recommended for publication by Editor Jessica Burgner-Kahrs upon evaluation of the Associate Editor and Reviewers' comments.
Kush Hari was supported by the NSF GRFP. The da Vinci Research Kit is supported by the National Science Foundation, via the National Robotics Initiative (NRI), as part of the collaborative research project “Software Framework for Research in SemiAutonomous Teleoperation” between The Johns Hopkins University (IIS1637789), Worcester Polytechnic Institute (IIS1637759), and the University of Washington (IIS 1637444). \textit{(Corresponding Author: Kush Hari)}} 
\thanks{$^{1}$Authors are with the Department of Electrical Engineering and Computer Science, UC Berkeley, California, USA
        {\tt\footnotesize \{kush\_hari,ziyang.chen,
        robotgksthf,goldberg\}@berkeley.edu}}%
\thanks{Digital Object Identifier (DOI): see top of this page.}
}
%
%

\markboth{IEEE Robotics and Automation Letters. Preprint Version. Accepted October, 2025}
{Hari \MakeLowercase{\textit{et al.}}: STITCH 2.0} 

%



\maketitle

\begin{abstract}
    \kushtimusprime{Surgical} suturing is a high-precision task \kushtimusprime{that impacts patient healing and scarring. Suturing skill varies widely between surgeons}, highlighting the need for robot assistance. Previous \kushtimusprime{robot} suturing works, such as STITCH 1.0 \cite{hari2024stitch}, struggle to fully close wounds due to inaccurate needle tracking and \kushtimusprime{poor thread management}. To address these challenges, we present STITCH 2.0, \kushtimusprime{an elevated augmented dexterity pipeline} with seven improvements including: improved EKF needle pose estimation, new thread untangling methods, and an automated 3D suture alignment algorithm. Experimental results over 15 trials find that STITCH 2.0 \kushtimusprime{on average} achieves 74.4\% wound closure with 4.87 sutures per trial, representing 66\% more sutures in 38\% less time compared to \kushtimusprime{the previous baseline}. When two human interventions are allowed, STITCH 2.0 averages six sutures with 100\% wound closure rate. 
\end{abstract}

\begin{IEEEkeywords}
Medical Robots and Systems, Surgical Robotics: Laparoscopy, Computer Vision for Medical Robotics
\end{IEEEkeywords}

%
\IEEEpeerreviewmaketitle

\input{Section/introduction}
\input{Section/related_work}
\input{Section/problem_statement}
\input{Section/method}
\input{Section/physical_experiments}
\input{Section/limitations}

\section*{Acknowledgment}

We thank Madison Veliky, Chung Min Kim, and Justin Kerr for their feedback on paper drafts and insightful conversations about the project. We thank Intuitive Surgical for the dVRK.

\ifCLASSOPTIONcaptionsoff
  \newpage
\fi



%
\bibliographystyle{IEEEtran}
\bibliography{main}

\end{document}

%% file: Section/introduction.tex
\section{Introduction}\label{sec:introduction}
\IEEEPARstart{S}{urgical} robots have revolutionized minimally invasive surgery, with Intuitive Surgical's da Vinci system performing over 2.6 million procedures in 2024 \cite{noauthor_2025-qu}. 
While these procedures require complete human control, recent advances in artificial intelligence (AI) present opportunities for surgical robot autonomy. However, the high-risk nature of surgery raises safety concerns for fully autonomous AI systems.

An alternative approach lies in augmenting surgeon capabilities by automating specific, well-defined subtasks. ``Augmented Dexterity" \kushtimusprime{proposed} by Goldberg and Guthart in 2024 \cite{goldberg2024augmented}, describes systems where surgical subtasks are performed by robots under close supervision of a human surgeon who is ready to take over at a moment's notice. \kushtimusprime{Augmented Dexterity has potential to increase surgical consistency and reduce tedium} amid a global \kushtimusprime{surgeon} shortage \cite{lawson2023global}.

One such repetitive subtask is suturing, where a surgeon uses a needle and thread to close a wound. Despite being common at the end of \kushtimusprime{almost every surgical }procedure, suturing quality varies \kushtimusprime{widely} between surgeons, especially when fatigue increases errors \cite{byrne2019surgical}. Uniform, \kushtimusprime{consistent} suturing \kushtimusprime{can reduce} infection and improve tissue healing \cite{goldberg2024augmented}.

\kush{While existing work has made great progress on components such as needle handover \cite{wilcox2022learning}, thread manipulation \cite{joglekar2025autonomous}, and suture placement \cite{bendikas2023learning}\cite{ramakrishnan2024automating}, these systems do not perform consecutive sutures. Previous multi-suture systems rely on either industrial robots using 3D tracking markers \cite{saeidi2022autonomous} or custom gripper attachments \cite{sen2016automating} not found in hospitals.} 

In contrast \kushtimusprime{to a previous paper}, STITCH 1.0 (Suture Throws Including Thread Coordination and Handoffs) \cite{hari2024stitch} uses the dVRK \cite{kazanzides2014open} \kushtimusprime{in an} augmented dexterity suturing pipeline including motion primitives for suture throws and needle handover based on a 6D needle pose estimation pipeline. \kush{However, STITCH 1.0 consistently faced challenges with needle dropping, thread tangling, and cross-stitching. These failures stemmed from noisy needle pose estimates, poor thread management, and uneven suture placement—all critical issues that must be addressed for clinical applications.}

\begin{figure}
    \centering
    \includegraphics[width=\linewidth]{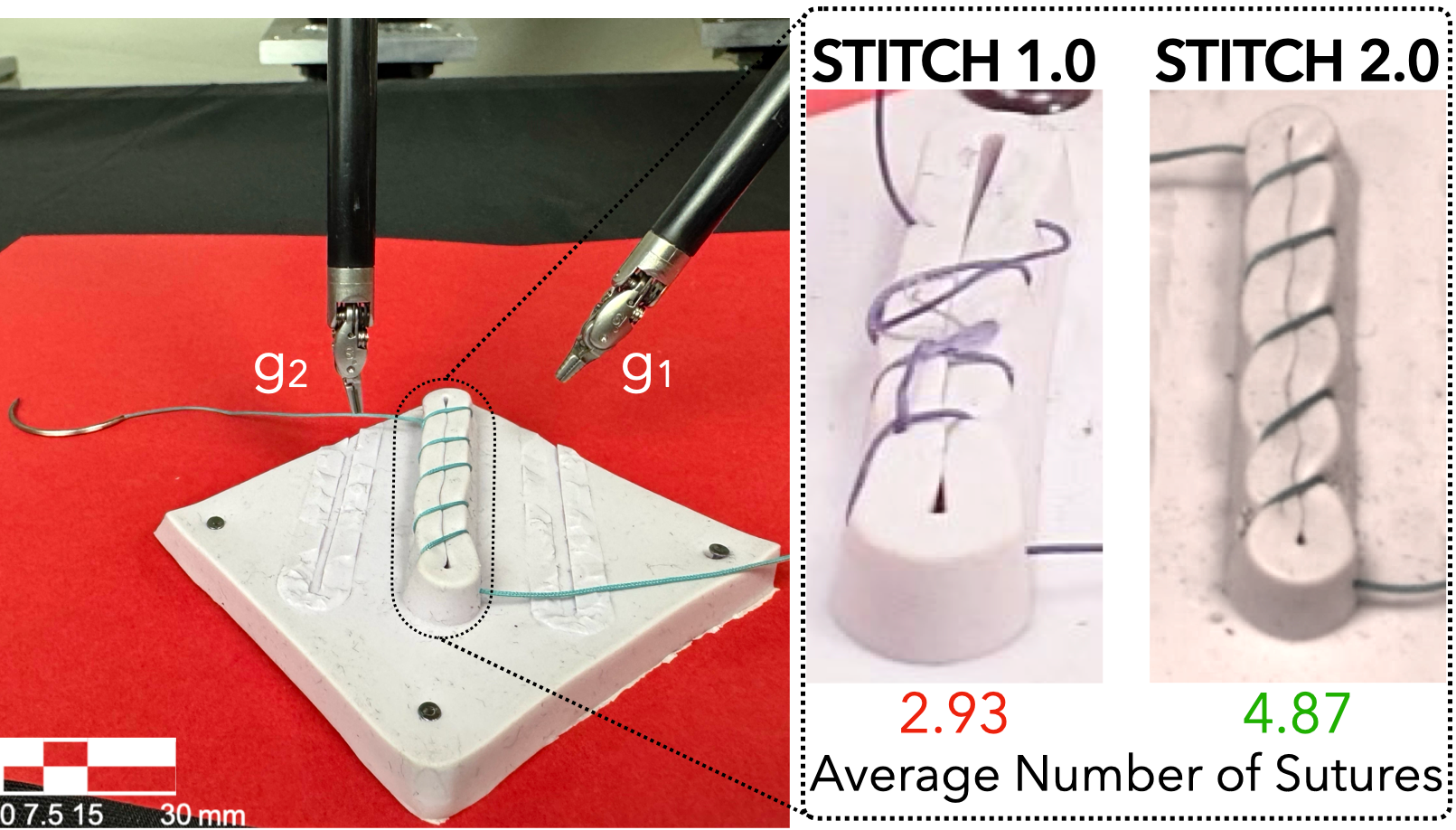}
    \vspace*{-6mm}
    \caption{\textbf{Augmented Suturing}: We use the da Vinci Research Kit, equipped with two grippers $g_{1}$ and $g_{2}$. \textbf{STITCH 1.0} averages 2.93 sutures \kushtimusprime{that were untightened and} did not fully close the wound. \textbf{STITCH 2.0} improved to 4.87 sutures, delivering tight, even sutures that fully close the wound.}
    \label{fig:splash_figure}
    \centering
    \vspace*{-7mm}
\end{figure}
\begin{figure*}[h]
    \centering
    \includegraphics[width=0.91\textwidth]{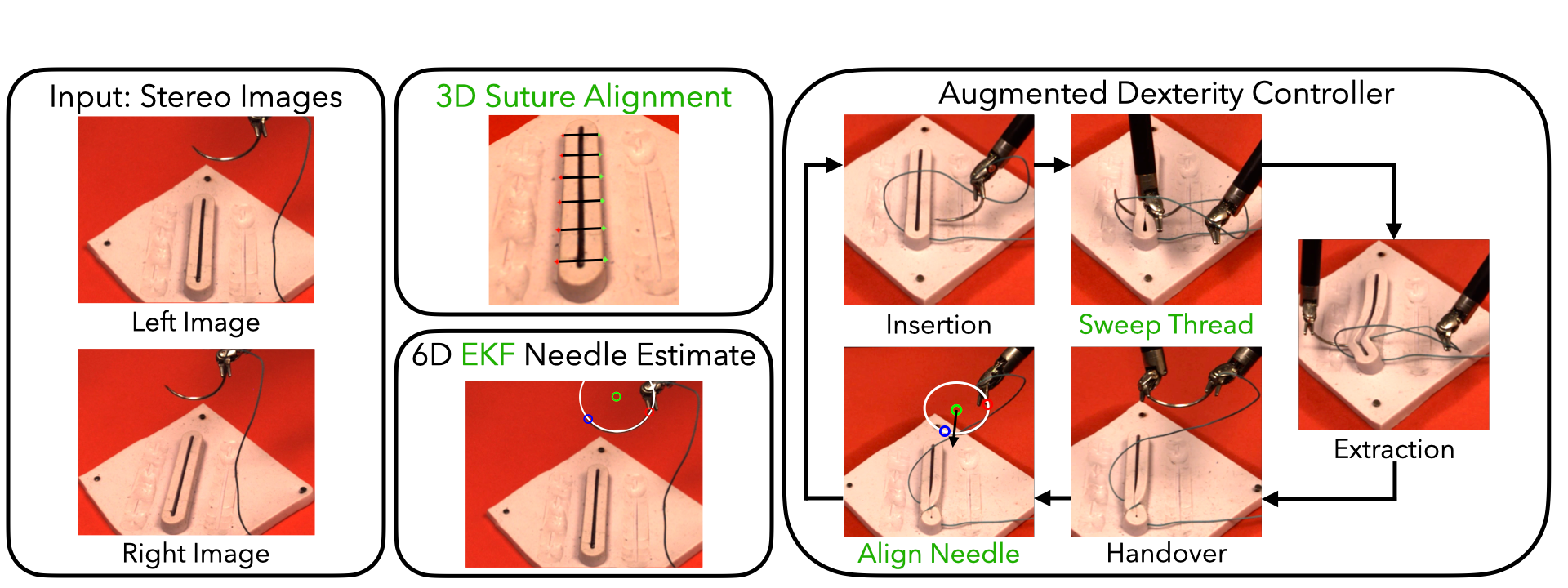}
    \vspace*{-3 mm}
    \caption{\textbf{Overview of the STITCH 2.0 pipeline with extensions from STITCH 1.0 \cite{hari2024stitch} in green.} STITCH 2.0 integrates a \textbf{3D Suture Alignment}, \textbf{6D EKF Needle Pose Estimator}, and \textbf{Augmented Dexterity controller}. The 3D suture alignment determines optimal insertion points, the 6D EKF Pose Estimator provides geometric data for precise needle handling, and the Augmented Dexterity controller executes suturing.}
    \label{fig:overall_pipeline}
    \vspace*{-6 mm}
\end{figure*}
To address these limitations, \kush{STITCH 2.0 \kushtimusprime{introduces} the following 7 components}:
\begin{enumerate} 
    \item Extended Kalman Filter integrated with needle pose pipeline to \kush{improve estimation accuracy}.
    \item \kush{Adaptive depth threshold to filter pointcloud noise around the tip point for accurate needle positioning} before insertion.
    \item 3D suture alignment to determine \kush{evenly-spaced} insertion and extraction points, eliminating the need for manual human input required by the STITCH 1.0 pipeline.
    \item Thread management methods that prevent thread tangling and cross-stitching \kush{to improve suture quality.}
    \item Needle-to-wound alignment system to improve suture placement accuracy.
    \item \kush{Flood-filled stereo images based on U-Net masks to reduce} needle pointcloud noise from light specularities.
    \item Skeletonized U-Net mask applied to depth image \kush{to eliminate needle boundary pointcloud noise}.
\end{enumerate}
\kushtimusprime{These improvements allow} STITCH 2.0 \kushtimusprime{to perform} 66\% more sutures in 38\% less time compared to STITCH 1.0.

%% file: Section/related_work.tex
\section{Related Work}\label{sec:related_studies}


\subsection{Needle Pose Estimation}
Accurate needle pose estimation is crucial for autonomous suturing \kush{to ensure precise suture placement and prevent tissue damage}. Previous approaches include segmentation-based methods \cite{hari2024stitch}\cite{wilcox2022learning}, which can struggle with noisy 3D reconstruction, along with end-to-end learning methods \cite{xu2024rnnpose}, which require detailed CAD models and separate training for different needle types. Alternatively, keypoint estimation methods, both in simulation \cite{jiang2023markerless} and in reality \cite{d2018automated}\cite{chiu2022markerless}, face challenges during needle manipulation. Although recent work \cite{chiu2023real} achieves submillimeter accuracy using gripper motion priors, tracking at high accuracy without such priors remains essential. Li's concurrent work \cite{Li2024monocular} on monocular keypoint estimation struggles with common suturing occlusions. Needle tracking in STITCH 2.0 improves on \cite{hari2024stitch} by using flood-filled stereo imaging to reduce the 3D reconstruction error common to segmentation-based methods.

\subsection{Suturing Subtask Automation}
The multistep nature of suturing has led researchers to focus on automating specific subtasks. These include needle handover \cite{wilcox2022learning}, suture placement \cite{bendikas2023learning}\cite{ramakrishnan2024automating}, thread manipulation \cite{joglekar2025autonomous}, and needle pickup \cite{d2018automated}\kush{, all with success rates exceeding 90\%. However, chaining these subtasks together into a full autonomous pipeline will accumulate error and lower the success rate for full wound suturing.}
\subsection{Suturing Automation Systems}
While some studies have explored suture trajectory planning \cite{marra2024mpc}, they primarily focus on simulation or avoid continuous suturing. Previous work on continuous suturing uses threadless needles or specialized equipment such as painted needles and gripper attachments \cite{sen2016automating}\cite{schwaner2021autonomous}. The Smart Tissue Autonomous Robot (STAR) system \cite{saeidi2022autonomous} demonstrates autonomous suturing in biological tissue but uses industrial robots and 3D tracking markers not available in surgical settings. 

\subsection{End-to-End Transformer Model for Surgical Tasks}
In recent work using a large transformer model, the Surgical Robot Transformer (SRT) \cite{kim2024surgical} employs imitation learning from demonstrations to perform isolated subtasks like needle handover and knot tying. \kush{SRT performs a single subtask while STITCH 2.0 performs 6 continuous suturing cycles for wound closure. Additionally, SRT uses four cameras (one stereo pair plus wrist-mounted cameras on each gripper), while STITCH 2.0 only uses one stereo pair.}

\begin{figure*}[t]
    \centering
    \includegraphics[width=0.95\textwidth]{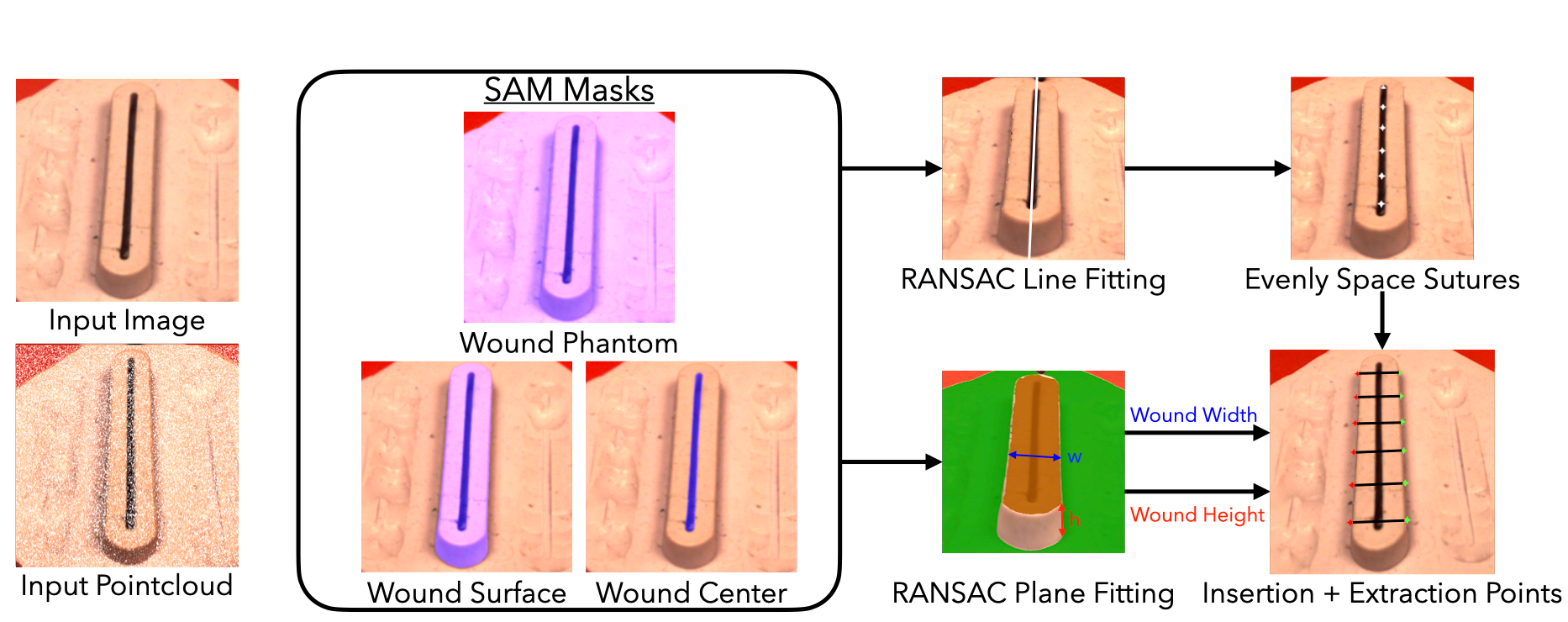}
    \vspace*{-4mm}
    \caption{The \textbf{3D Suture Alignment} starts with generating a scene pointcloud using RAFT-Stereo \cite{lipson2021raft}, followed by segmentation of the wound center, surface, and phantom with Segment Anything (SAM) \cite{kirillov2023segment}. The wound height is calculated by measuring the distance between the wound and phantom surfaces through RANSAC fitting. A 3D line representing the wound center is obtained by projecting points onto the top wound surface plane and fitting with RANSAC. Suture positions are evenly distributed along the centerline, and the wound width and height are used to determine the insertion points for needle placement.}
    \label{fig:3d_suture_placement}
    \vspace*{-6mm}
\end{figure*}

%% file: Section/problem_statement.tex
\section{Problem Statement}\label{sec:problem_definition}
We consider a laparoscopic surgery setting with a raised wound phantom and bimanual classic dVRK Intuitive Surgical robot (Fig. \ref{fig:splash_figure}). Given a stereo camera video stream, the objective is to evenly suture the wound closed by planning suitable needle entry and exit points and manipulating a surgical needle with thread.

We make the following assumptions:
\begin{enumerate}
    \item The needle is a semi-circle with known radius\kush{, common for clinical suturing}.
    \item There is sufficient thread attached to the needle to fully suture the wound.
    \item \kushtimusprime{The thread starts untangled at the beginning of each experiment.}
    \item \kush{We use }a raised linear wound phantom, \kush{common for resident training \cite{suturepad}. The wound is placed orthogonal to the stereo camera pair for easier suture alignment. See Limitations and Future Work section for more challenging wound configurations.}

    \item The stereo camera pair is calibrated and its \kush{intrinsic and extrinsic} parameters are known.
    \item The dVRK is calibrated with an RNN based on \cite{hwang2020efficiently} to reduce kinematic errors.
\end{enumerate}

We evaluate STITCH 2.0 based on 3 evaluation metrics: completion time in seconds, number of successful sutures, and percentage of wound gap closed. A successful suture throw is defined by passing the needle through the wound at the desired insertion/extraction points and performing sufficient suture cinching without thread over-tightening, thread tangling, or cross-stitching. Percentage of wound gap closed is defined as the number of stitches that closed the wound divided by the total number of planned stitches.

%% file: Section/method.tex
\section{Methods}
\label{sec:method}
STITCH 2.0 \kushtimusprime{iterates over} 3 phases as specified in Fig. \ref{fig:overall_pipeline}.
\begin{enumerate}[label=\Alph*.]
    \item \textbf{3D Suture Alignment} autonomously determines the optimal insertion/extraction points for each suture based on stereo imaging, improving on previous work \cite{hari2024stitch} where the insertion/extraction points were manually provided.
    \item \textbf{6D Needle Pose Estimator with EKF}.
    \item \textbf{Augmented Dexterity Controller} manages gripper movements to complete suture throws, needle extraction, needle handover, and thread slack management based on perception inputs from \kushtimusprime{Phases A and B}.
\end{enumerate}

\subsection{3D Suture Alignment}
Uniform suture alignment can facilitate patient healing and reduce tissue scarring. It is important to properly space sutures because insufficient spacing can restrict tissue oxygenation, while excessive spacing can fail to provide adequate closure force \cite{barnes2013suture}. \kush{While STITCH 1.0 required user-specified insertion and extraction points, STITCH 2.0 performs autonomous suture alignment (Fig. \ref{fig:3d_suture_placement}).}

\kushtimusprime{The system starts} by building a 3D wound model. \kushtimusprime{It creates} a scene pointcloud by passing stereo images into RAFT-Stereo \cite{lipson2021raft}, an off-the-shelf stereo network. \kushtimusprime{It} then employs Segment Anything (SAM) \cite{kirillov2023segment} to segment the wound center, surface, and phantom, creating separate pointclouds for each component. Because we assume \kush{the wound is flat and raised}, \kushtimusprime{the system calculates} the height by measuring the distance between the top wound surface plane and the bottom phantom surface plane. Both planes are determined through Random Sample Consensus (RANSAC) fitting of their respective pointclouds.

To determine the suture positions, STITCH 2.0 projects the wound center pointcloud to the top wound surface plane and fits a center line using RANSAC. Based on the force closure model from \cite{ramakrishnan2024automating}, \kushtimusprime{the robot attempts to} evenly distribute 6 suture positions along this centerline.

To find the 3D \kushtimusprime{needle} insertion-extraction points, \kushtimusprime{the system} first finds the wound width vector. The direction is computed as the cross product of the wound surface normal and centerline. The magnitude is calculated by projecting the wound surface pointcloud onto this direction unit vector and finding the maximum distance between projected points. With the wound width and wound height, \kushtimusprime{the system} calculates the 3D insertion-extraction points by offsetting the \kush{centered suture positions} by half the wound width and half the wound height \kush{in their respective directions}, so the needle insertion-extraction points are at mid-phantom height \kush{on the wound boundary}.

\begin{figure*}[t]
    \centering
    \includegraphics[width=0.95\textwidth]{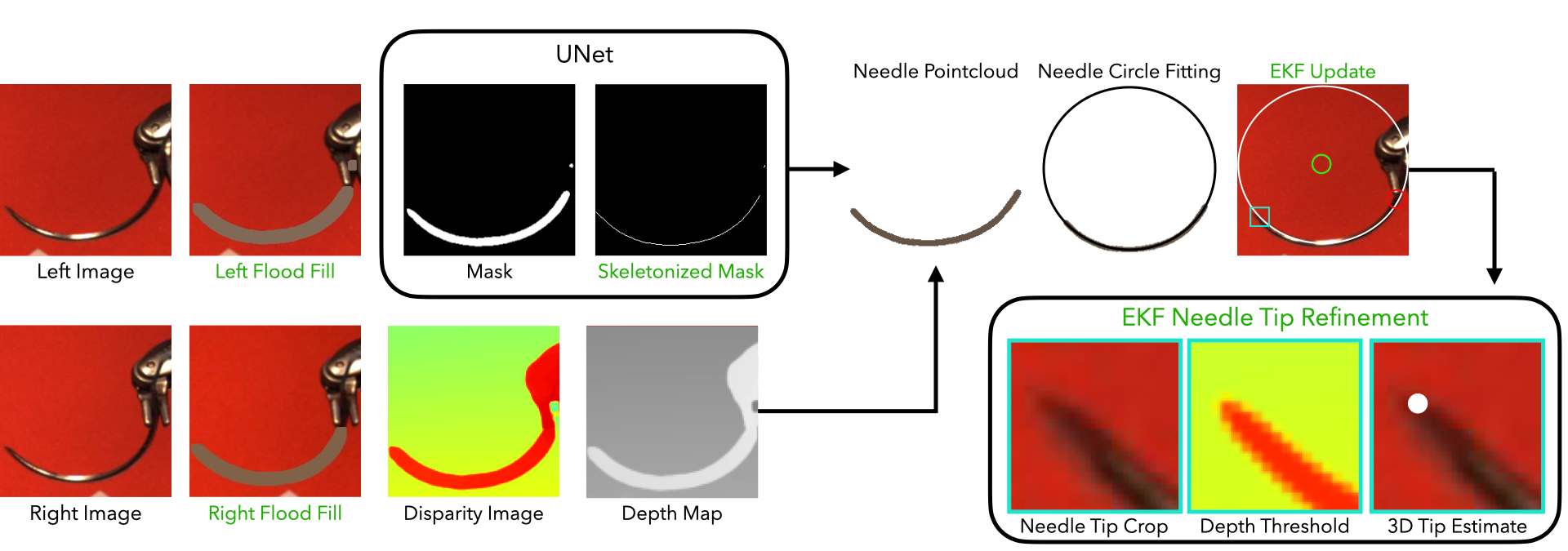}
    \vspace*{-4mm}
    \caption{The \textbf{6D Needle Pose Estimator} begins by training a U-Net for needle segmentation. A flood-fill approach is applied to reduce noise and eliminate specularities before processing through RAFT-Stereo \cite{lipson2021raft} to generate a disparity image. The disparity image is converted into a depth map, and a skeletonized mask is used to improve pose estimates and accelerate circle fitting. The pose is further stabilized with an Extended Kalman Filter (EKF). Needle tip estimation is refined by adaptive thresholding and computing the intersection of the pixel ray with the EKF 3D circle estimate, enabling precise needle manipulation for insertion and suture alignment. \textbf{New contributions with STITCH 2.0 from STITCH 1.0 \cite{hari2024stitch} are in green.}}
    \label{needle_estimation_pipeline}
    \vspace*{-6 mm}
\end{figure*}

\subsection{6D Needle Pose Estimator with EKF}
The STITCH 2.0 6D pose estimator with EKF determines the best-fit 3D circle for the surgical needle, providing the circle center coordinates, needle endpoint positions, and the normal vector of the needle plane as outlined in Fig. \ref{needle_estimation_pipeline}. Previous realtime needle pose estimation methods \cite{chiu2022markerless}\cite{chiu2023real}\cite{Li2024monocular} use keypoints and struggle in common suturing scenarios where many keypoints on the needle can be occluded by the grippers or the wound tissue.

We begin by training a U-Net \cite{ronneberger2015u} for needle segmentation. The U-Net is trained on 1200 stereo image pairs \kush{showing the needle in random poses and lighting conditions. The number of training images was iteratively increased until the U-Net achieved robust performance across image conditions typical in suturing, including scenarios with both needle and its shadow present.} \kush{Training data is collected by painting the needle with ultraviolet (UV) paint and performing color segmentation on UV images to generate ground truth needle masks. \cite{thananjeyan2022all}.} However, UV paint degradation can result in incomplete masks, so we use SAM \cite{kirillov2023segment} to generate complete ground truth masks for the affected images.

\kushtimusprime{During suturing, the system uses} the U-Net to segment the needle in both the left and right images and flood-fill the needle regions with the average color. This is significant because in STITCH 1.0, the stereo pointcloud was noisier in areas where the needle had white spots caused by the needle’s reflective surface. Replacing the entire needle region with a uniform average color in the left and right images eliminates these visual specularities. Therefore, \kushtimusprime{the system passes} the flood-filled left and right images into RAFT-Stereo \cite{lipson2021raft} to generate a disparity image and depth map.

From there, \kushtimusprime{the system applies} a skeletonized U-Net mask to the depth map and creates the needle pointcloud. In STITCH 1.0 \cite{hari2024stitch}, we applied the unmodified U-Net mask to the depth map, and observed the needle pointcloud was noisiest at the boundaries. \kushtimusprime{Here, the system uses} Zhang's method \cite{zhang1984fast} to iteratively remove border pixels and create a single pixel-width skeletonized mask. \kush{Because we use the skeletonized mask to obtain the needle pointcloud, STITCH 2.0 yields robust needle pose estimates while simultaneously accelerating the subsequent circle fitting by 2-3x by processing just a fraction of the original pointcloud.}

\kush{After these filtering steps, the pointcloud is stable enough to perform the 3D circle fitting process from STITCH 1.0 \cite{hari2024stitch}. This involves RANSAC plane fitting to remove lingering pointcloud outliers and estimate the needle normal vector. The algorithm iteratively samples random subsets of the needle pointcloud, fits a plane to each subset, and selects the subset with the most inliers for the final plane fit. The RANSAC inliers are then projected onto the best-fit plane to optimize circle parameters, with the two farthest points selected as needle endpoints.}

\begin{figure*}[t]
    \centering
    \includegraphics[width=\textwidth]{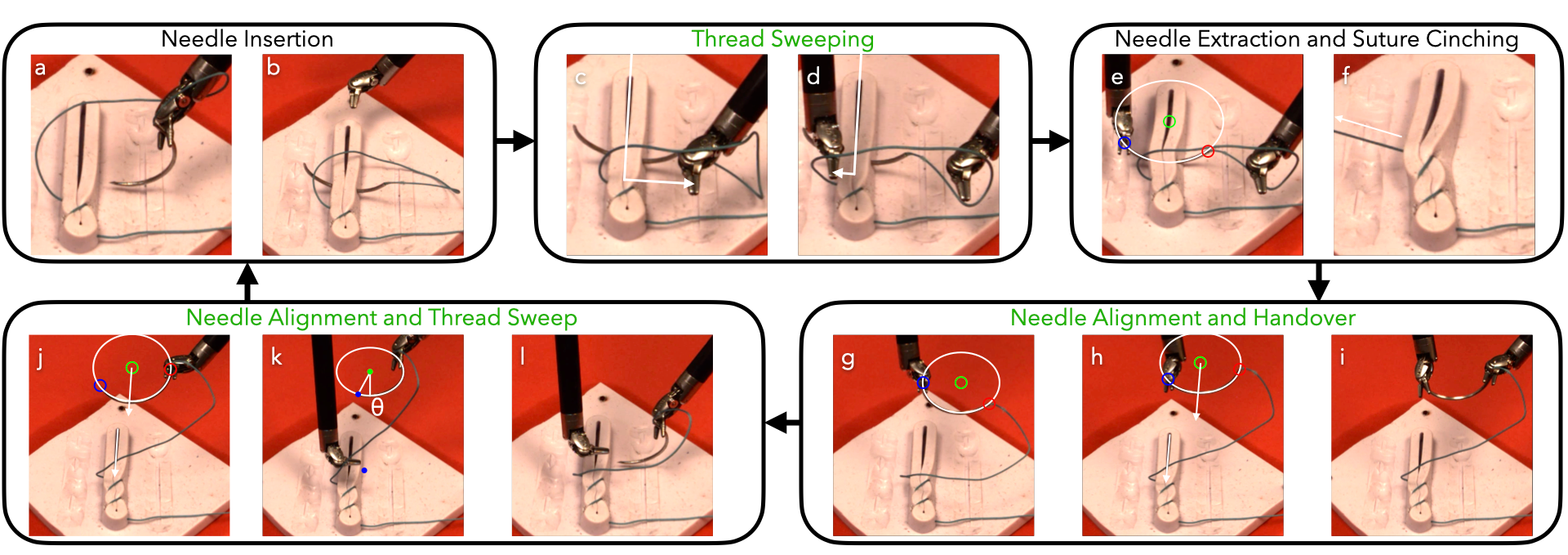}
    \vspace*{-9 mm}
    \caption{The \textbf{Augmented Dexterity Controller} enables suturing automation through five components: (a), (b) needle insertion, (c), (d) thread sweeping, (e), (f) needle extraction with thread cinching, (g)-(i) needle alignment and handover, and (j)-(l) pre-insertion alignment. Needle insertion involves precise suture positioning. Thread sweeping prevents tangling, while extraction involves cinching for wound closure. Needle alignment and handover correct needle orientation, and pre-insertion alignment facilitates a high-quality insertion. \textbf{New STITCH 2.0 contributions from STITCH 1.0 \cite{hari2024stitch} are in green.}}
    \label{motion_controller}
    \vspace*{-5mm}
\end{figure*}

\begin{figure}[t]
    \centering
    \includegraphics[width=0.85\linewidth]{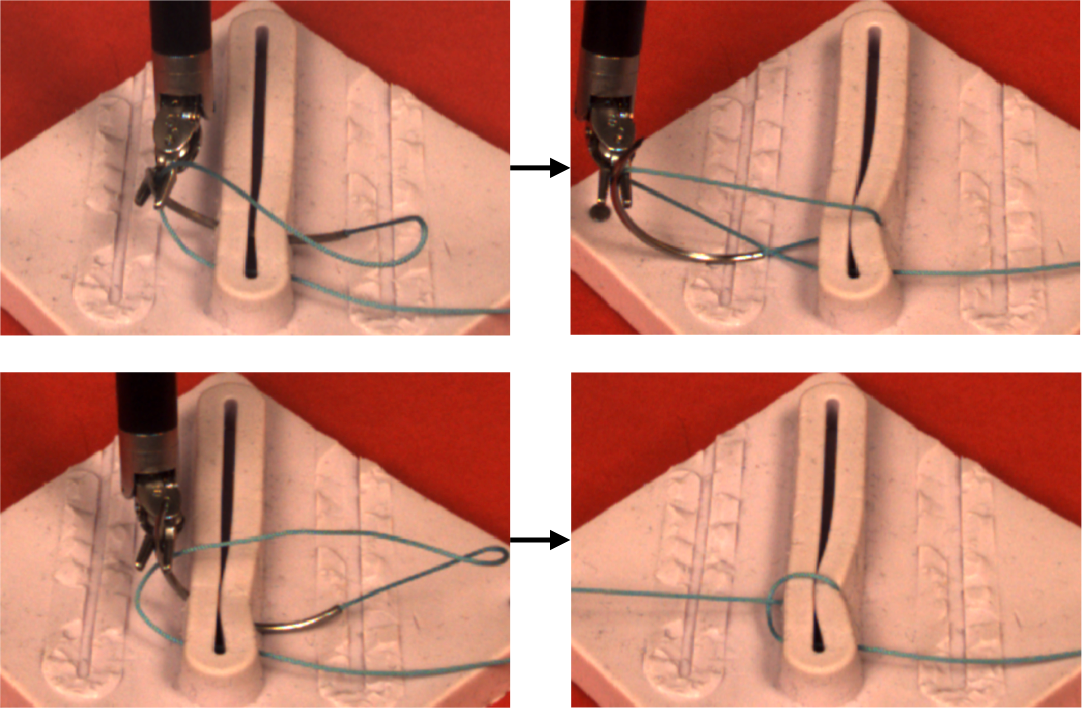}
    \vspace*{-3 mm}
    \caption{\textbf{Previous Thread Tangling Failures.} (top) Gripper grasps thread and needle together leading to thread tangling. (bottom) Gripper grasps needle while thread is behind the needle leading to cross-stitch.}
    \label{fig:thread_failure}
    \vspace*{-6.4mm}
\end{figure}
\subsubsection*{EKF Needle Estimate}
\kush{While circle fitting provides reliable estimates for the needle normal vector and center point, the endpoints have up to 3-5 mm error due to noisy pointcloud data and U-Net segmentation errors at the endpoints. This inaccuracy results in poor needle insertions and sutures.} 

To address this in STITCH 2.0, \kushtimusprime{the system generates} a new Extended Kalman Filter (EKF) before each needle interaction step (needle extraction, handover, alignment, etc.) when the needle is stationary. We do this as opposed to having a single EKF track the needle throughout the procedure because the needle estimates are noisiest during needle motion and STITCH 2.0 only requires needle pose estimation \kush{directly} before needle interaction steps\kush{, as described in the Augmented Dexterity Controller section (Sec. C)}. We use EKFs as opposed to standard Kalman Filters because EKFs can handle non-Gaussian noise by performing linearization around the current needle state mean and covariance.

\kush{The EKF needle state is a 13-dimensional vector comprising the needle center point $(c_x,c_y,c_z)$, left endpoint $(l_x,l_y,l_z)$, right endpoint $(r_x,r_y,r_z)$, normal vector $(n_x,n_y,n_z)$, and radius $r$:
$$\mathbf{x} = [c_x, c_y, c_z, l_x, l_y, l_z, r_x, r_y, r_z, n_x, n_y, n_z, r]^T$$
Since the measurement vector has identical structure to the state vector, the measurement model uses the identity matrix: $\mathbf{H} = \mathbf{I}_{13}$. Because the needle remains stationary between EKF steps, the state transition model is also the identity matrix: $\mathbf{F} = \mathbf{I}_{13}$. These state and measurement models can now be directly applied to the EKF.}  

For each EKF, the initial state and covariance is determined by averaging the first seven measurements. Then, \kushtimusprime{the system applies} the EKF update step to the next three measurements \kush{that are within 3 standard deviations of the initial state} to get the final needle estimate. \kushtimusprime{The number of initial and update measurements are empirically determined by iterating from zero until the estimates consistently converge.}

With this stable needle pose estimate from the EKF, \kushtimusprime{the robot} can now accurately determine the needle tip point. We first crop a 200x200 patch centered around the EKF-estimated needle tip in the disparity image (Fig. \ref{needle_estimation_pipeline}). Then, we perform contour detection \cite{zhang1984fast} with adaptive depth thresholding \kush{to create a skeletonized needle mask in the image crop. Since this mask maintains a consistent 1-pixel width, the end corner of the mask corresponds to the needle tip in pixel space.} While the disparity image signal is strong enough for an adaptive threshold, directly projecting that pixel to a 3D point is too noisy. Since the EKF merges prior measurements resulting in a stable needle circle estimate, \kushtimusprime{the system determines the depth by deprojecting the needle tip pixel to a 3D ray and computing the intersection between the ray and the EKF 3D circle estimate.} \kush{This provides a reliable 3D estimate of the needle tip, enabling the augmented dexterity controller to maintain proper needle alignment with the wound during suturing, as described in the next section.}

\subsection{Augmented Dexterity Controller}\label{sec:motion_controller}

The augmented dexterity controller implements suturing automation \kushtimusprime{through the two dVRK grippers ($g_{1}$ and $g_{2}$)} with five key subtasks (Fig. \ref{motion_controller}): needle insertion, thread sweeping, needle extraction with thread cinching, needle alignment and handover, and pre-insertion needle alignment. \kush{Because these subtasks are negatively affected by the inaccurate cable-driven kinematics,} we perform an RNN calibration \cite{hwang2020efficiently} for each dVRK gripper (Fig. \ref{fig:splash_figure}).

\subsubsection{Surgical Needle Insertion}
The insertion process begins with positioning the needle tip at the planned insertion point (Fig. \ref{motion_controller}(a)). \kushtimusprime{Gripper} $g_{1}$ executes a 3-step twist motion \kushtimusprime{that reduces} strain on the wound phantom as the needle is inserted. \kushtimusprime{This motion was empirically determined so that $g_{2}$ can grasp the needle for extraction.} \kush{Initially, the needle is translated into the wound by one wound width \kushtimusprime{(9 mm)}, ensuring the needle tip completely penetrates through the wound. The needle is then rotated 40 degrees around the wound center line, leveraging the circular needle geometry. Finally, the needle undergoes an additional 2 mm translation to prevent the needle from slipping out the wound.} Afterward, $g_{1}$ releases the needle when insertion is completed (Fig. \ref{motion_controller}(b)).

\subsubsection{Thread Slack Management} 
Poor thread management can cause thread tangling (Fig. \ref{fig:thread_failure} top), or ``cross-stitching" that prevents proper suture tightening and wound closure (Fig. \ref{fig:thread_failure} bottom). Both of these issues occur when the thread rests behind the needle during extraction \kushtimusprime{as seen in Fig. \ref{fig:thread_failure}}. In STITCH 1.0 \cite{hari2024stitch}, thread management involves just moving $g_{2}$ down the center of the wound to push the thread forward. However, the thread often springs back \kush{behind the needle} leading to thread tangling errors. For STITCH 2.0, we empirically determine a trajectory where $g_{1}$ and $g_{2}$ coordinate to ensure the thread remains in front of the needle: $g_{1}$ moves down the wound center, lifts the thread, and moves to the right (Fig. \ref{motion_controller}(c)). This provides ample clearance for $g_{2}$ to move down the wound center and lift the thread on the other side to ensure the thread stays in front of the needle (Fig. \ref{motion_controller}(d)). Then, $g_{2}$ returns to the home position, but $g_{1}$ maintains its position \kush{to hold the thread} during needle extraction. \kush{This move prevents thread tangling across various configurations (Fig. \ref{fig:thread_failure}). When the thread is already positioned in front of the needle, the move does not disrupt this configuration, allowing the thread to remain untangled throughout the extraction process.}
\input{Tables/ablation_table}
\begin{figure}[t]
    \centering
    \includegraphics[width=0.9\linewidth]{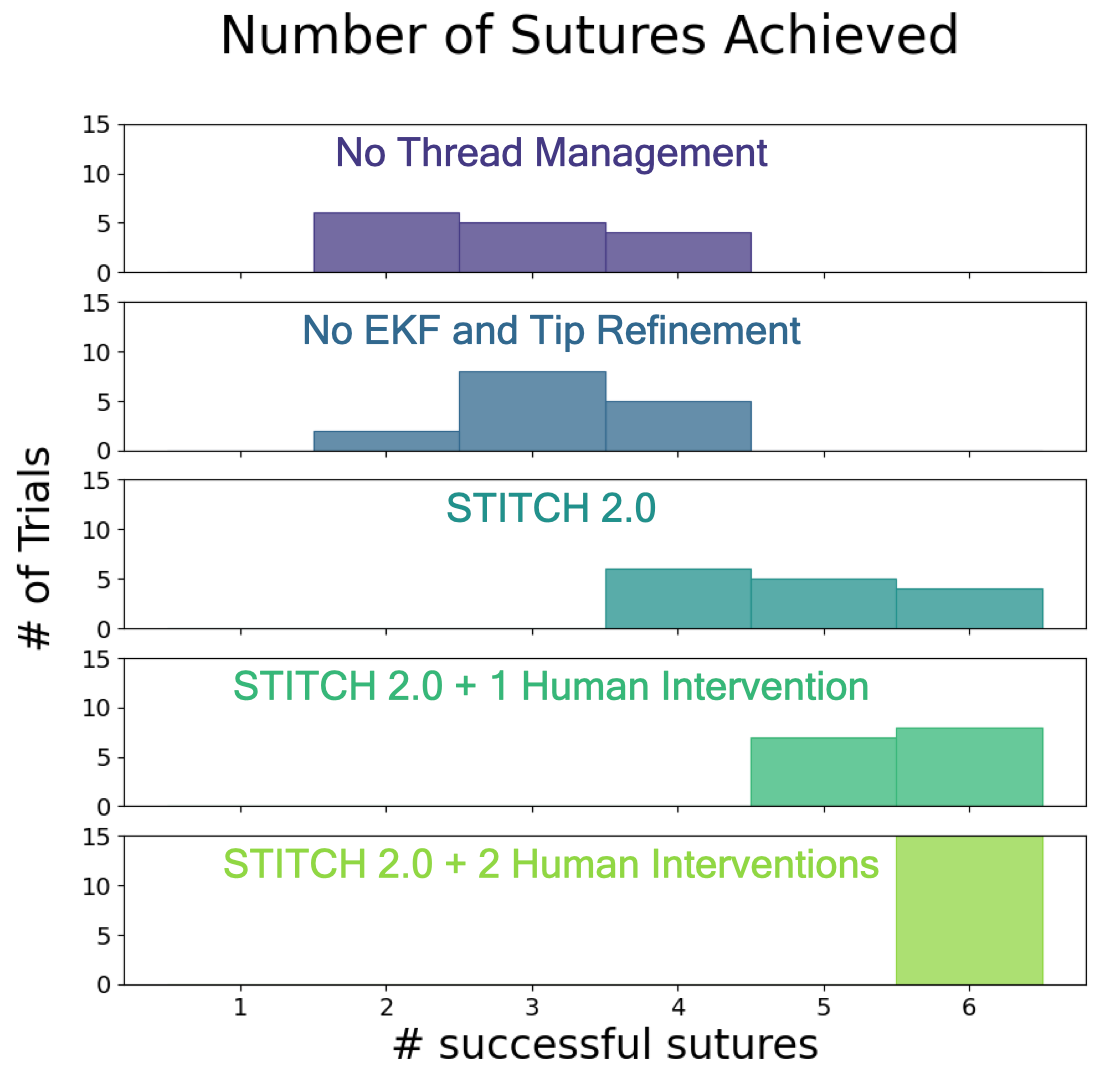}
    \vspace*{-0.15in}
    \caption{\textbf{Histogram of the Number of Sutures Achieved by Method.} The results in the histogram shown above is for 15 trials of 6 sutures per each of the 5 methods in the ablation up to 450 suture trials.}
    \label{fig:results}
    \vspace*{-0.3in}
 \end{figure}
\subsubsection{Surgical Needle Extraction with Thread Cinching}
Extraction begins with $g_{2}$ grasping the needle 2 mm from its tip (based on the 6D pose estimate) to avoid dulling the needle tip (Fig. \ref{motion_controller}(e)). Once grasped, the needle is twisted to extract it out mirroring the earlier insertion twist motion. 

Subsequently, the surgical thread is pulled to close the wound and then the needle is moved to the handover position (Fig. \ref{motion_controller}(f)). The following equation is used to determine the appropriate \kushtimusprime{goal point} to pull the thread for cinching:
\begin{equation} D_{i} = [n * d - ((i - 1) * d)] 
  \label{eqn_cinching_motion} \end{equation}
Where $D_{i}$ is the thread cinching translation, $n$ is the number of planned sutures, $i$ is the current suture number, and $d$ is the per-suture thread length determined by the 3D suture alignment.

\subsubsection{Needle Alignment and Handover}
A common failure case \kush{for needle handover} in STITCH 1.0 \cite{hari2024stitch} is that the needle snapped to an orientation \kush{where it was misaligned with the wound} when $g_{2}$ released the needle. To mitigate this, \kushtimusprime{STITCH 2.0 rotates the needle such that the needle normal vector is parallel to the wound allowing for orthogonal tissue entry (Fig. \ref{motion_controller}(g). Then, STITCH 2.0 rotates the needle around the normal until both endpoints are level (Fig. \ref{motion_controller} (h)). Leveling the needle endpoints is crucial for preventing needle deflection and misalignment after gripper release by ensuring $g_{1}$ has the same approach angle as $g_{2}$.} With the needle properly oriented, gripper $g_{1}$ grasps 2 mm below the estimated needle endpoint connected to the thread (Fig. \ref{motion_controller}(i)). Once $g_{1}$ and $g_{2}$ have grasped the needle, $g_{1}$ releases the needle \kush{and it remains at the same orientation.}

\begin{figure}[t]
    \centering
    \vspace*{0.025in}
    \includegraphics[width=0.95\linewidth]{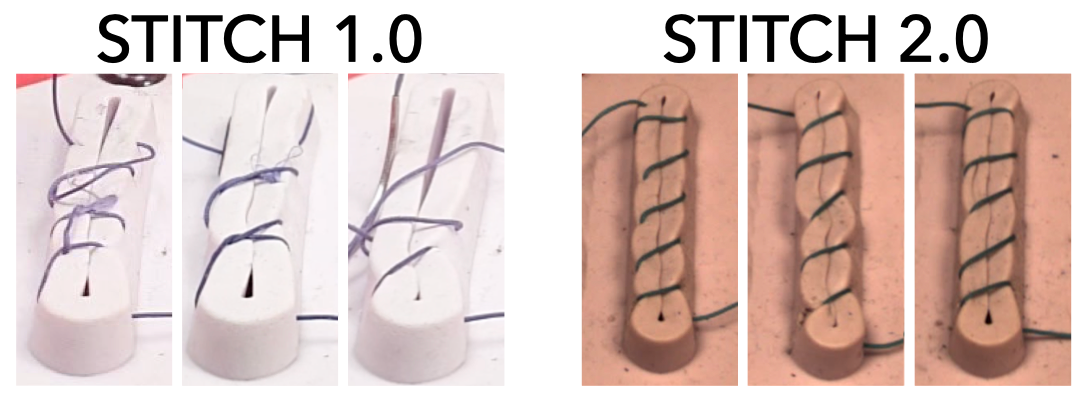}
    \vspace*{-0.1in}
    \caption{\kush{\textbf{Qualitative Comparison.} While STITCH 1.0 (left) produces messy sutures that are insufficient for wound closure, STITCH 2.0 (right) delivers uniform sutures that can achieve successful closure.}}
    \label{fig:stitch_comparisons}
    \vspace*{-0.2in}
\end{figure}

\subsubsection{Pre-Insertion Needle Alignment}
\kushtimusprime{To ensure the needle enters the tissue orthogonally for a high quality stitch and minimal tissue damage, the system first rotates the needle so its normal is aligned to the wound similar to the previous step.} (Fig. \ref{motion_controller}(j)). The needle is then rotated around its normal vector to establish a 20-degree angle between the vertical axis and the needle tip-to-center line (Fig. \ref{motion_controller}(k)). \kush{This empirically-determined angle improves insertion quality by leveraging the flat, linear tip design, which translates smoothly into tissue when positioned at this orientation.} With the needle in this configuration, $g_{2}$ then pushes the thread and holds it forward while $g_{1}$ moves to the next insertion point (Fig. \ref{motion_controller}(l)). Overall, this process iterates until all planned sutures are completed.

%% file: Tables/ablation_table.tex
\begin{table*}[htbp!]
    \vspace{0.25in}
    \caption{\label{ablation-table}STITCH 2.0 with Ablations for 15 Trials of 6 sutures for each method (Maximum 450 Suture Trials).
    }
    \centering
    \resizebox{\textwidth}{!}{%
    \begin{tabular}{l|c|rrrr rrcc r r}
    \toprule[1pt]
    \multicolumn{1}{c|}{\multirow{2}{*}{}} & \textbf{Avg. Number} & \textbf{Single-Suture} & \textbf{Wound Gap} & \textbf{Mean Time} & \multicolumn{5}{c}{\textbf{Error types}} & \textbf{Needle Estimate} \\
     & \textbf{of Sutures} & \textbf{Success Rate (\%)} & \textbf{Closure Rate (\%)} & \textbf{per Suture (s)} & A & T & I & M & \textbf{Total} & \textbf{Success Rate (\%)}\\
    \midrule[0.1pt]
    \textbf{\kush{STITCH 1.0 \cite{hari2024stitch}}} & \kush{2.93 ± 0.70} & \kush{69.4} & \kush{22.3} & \kush{159.3} & \kush{0} & \kush{2} & \kush{8} & \kush{5} & \kush{15} & \kush{42}\\
    \textbf{\kush{STITCH 1.0 \cite{hari2024stitch} + 2 Human Corr.}} & \kush{4.47 ± 0.99} & \kush{83.3} & \kush{35.3} & \kush{141.9} & \kush{0} & \kush{0} & \kush{8} & \kush{5} & \kush{13} & \kush{42}\\
    \textbf{STITCH 2.0 w/o Thread Mgmt.} & 2.87 ± 0.83 & 74.1 & 
    39.4 & 81.02 & 2 & 7 & 2 & 4 & 15 & -\\
    \textbf{STITCH 2.0 w/o EKF} & 3.20 ± 0.68 & 76.2 & 44.5 & \textbf{61.60} & 2 & 4 & 3 & 6 & 15 & 85.98\\
    \textbf{STITCH 2.0} & \textbf{4.87 ± 0.83} & \textbf{86.9} & \textbf{74.4} & 98.74 & 3 & 3 & 5 & 0 & \textbf{11} & \textbf{97.56}\\
    \midrule[0.5pt]
    \textbf{STITCH 2.0 + 1 Human Corr.} & 5.53 ± 0.52& 92.2 & 87.8 & 103.40 & 2 & 2 & 1 & 0 & 5 & \textbf{97.56}\\
    \textbf{STITCH 2.0 + 2 Human Corr.} & \textbf{6.00 ± 0.00} & \textbf{100.0} & \textbf{100.0} & 101.60 & 0 & 0 & 0 & 0 & \textbf{0} & \textbf{97.56}\\
    \bottomrule[1pt]
    \end{tabular}%
    }\\[0.05in]
    \footnotesize
    \raggedright
    \kush{\textbf{STITCH 2.0 Comparisons (N = 15): 
    } We evaluate each method on 15 trials of 6 sutures each. STITCH 2.0 executes more sutures than both ablations, STITCH 1.0, and STITCH 1.0 with human corrections. Furthermore, allowing for up to 2 human corrections led to 100\% wound closure rate with 0 errors as seen in the bottom row of the table. See the Suturing Experiments section for more details about the different error types. }
    \vspace{-20pt}    
\end{table*}

%% file: Section/physical_experiments.tex
\section{Physical Experiments}\label{sec:experiment}

\kush{Experiments evaluate (1) the ability of STITCH 2.0 to suture a wound closed compared to STITCH 1.0, (2) the impact of the EKF and thread management on STITCH 2.0, (3) the impact of augmented dexterity on STITCH 2.0 by allowing for minimal human corrections, and (4) the needle pose estimation accuracy across all suturing trials.}

For robot execution, we use a bimanual classic dVRK \cite{kazanzides2014open}. The wound phantom is a soft tissue suture pad \cite{suturepad} used for medical education and training. We use a pair of fixed RGB stereo cameras (Prosilica GC 1290, Allied Vision). The stereo cameras record 1280x960 images at 33 frames per second. We use variable-length (30-75 cm) green braided  $\text{Ethibond Excel}^{\text{TM}}$ suture thread. Attached to the thread is a V-37 40mm half-circular tapercut surgical needle \cite{ethibond}.

\begin{figure}[t]
    \centering
    \includegraphics[width=0.9\linewidth]{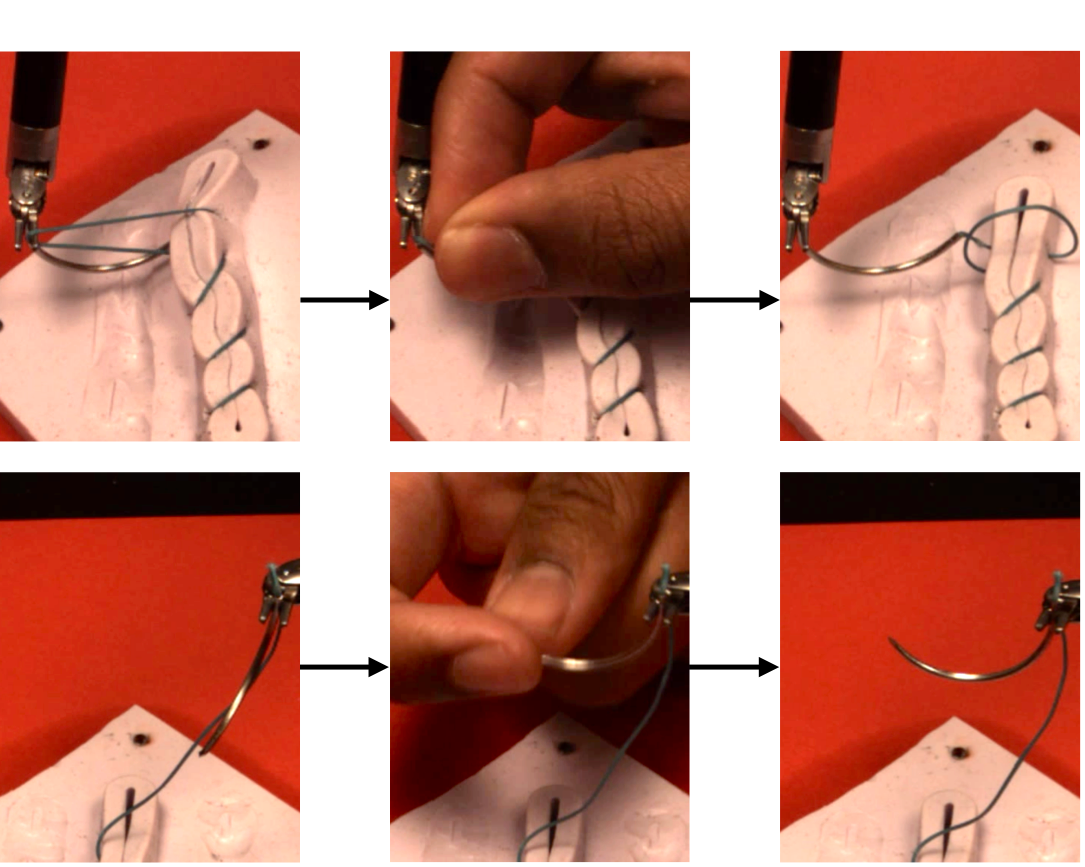}
    \caption{\textbf{Sample Human Interventions for STITCH 2.0.} (top) Human removes thread from gripper to prevent tangling during extraction. (bottom) Human corrects needle orientation after handover.}
    \label{fig:human_intervention}
    \vspace*{-6mm}
 \end{figure}
\subsection{Suturing Experiments}
We evaluated STITCH 2.0 across 15 \kushtimusprime{experiments} of six \kushtimusprime{consecutive suture trials} each. A suture is successful if the needle \kushtimusprime{is} inserted and extracted from the phantom without damaging the wound and the \kushtimusprime{suture thread is correctly tightened across the wound for closure}. We measure mean sutures achieved, single-suture throw success rate, wound gap closure rate, mean time per suture, \kushtimusprime{and needle estimate success rate}. During experiments, we observed 4 different types of failures: (A)lignment error where \kush{the needle alignment step failed, leaving the needle unaligned to the wound causing a failed insertion, extraction, or handover}, (T)hread management error where the thread gets tangled with the robot or is improperly cinched over the wound leading to a low quality suture, (I)nsertion error where the needle misses the phantom or is inserted in the phantom such that it cannot be extracted, and (M)issed grasp where an inaccurate needle pose estimate leads to the gripper missing a grasp during either extraction or handover. 

Table \ref{ablation-table} and Fig. \ref{fig:results} display experimental results. STITCH 2.0 achieves an 86.9\% single-suture success rate and 74.4\% wound gap closure rate. Additionally, STITCH 2.0 achieves an average of 4.87 sutures at 98.74 seconds per suture—a 66\% increase in higher-quality sutures and 38\% decrease in time compared to STITCH 1.0 \kush{(Fig. \ref{fig:stitch_comparisons})}. \kush{Furthermore, STITCH 2.0 averages 6 sutures with 2 human interventions (See Augmented Dexterity in Experiments for more details).} 


\subsection{Ablations}

We compare STITCH 2.0 against two \kush{ablation} configurations: one without thread management and one without EKF and needle tip refinement (Table \ref{ablation-table}).

STITCH 2.0 \kushtimusprime{completes} 2.00 additional sutures \kushtimusprime{on average} compared to the ablation without thread management resulting in a 35\% increase in wound gap closure. This improvement comes from the thread management method that reduces thread tangling by 57\% despite taking 17.72 more seconds.

Similarly, STITCH 2.0 averages 1.67 additional sutures compared to the ablation without the EKF and needle tip refinement. While this additional EKF and needle tip refinement step averages an extra 37.14 seconds, it reduces alignment and missed grasp failures by 50\%.

\kush{Because STITCH 2.0 has a higher average suture count compared to its ablations, it has increased error counts in certain individual categories. However, STITCH 2.0 still achieves fewer total errors across all categories combined, as demonstrated in the Total Error column of Table \ref{ablation-table}.}

\subsection{Augmented Dexterity in Experiments}
In the context of Augmented Dexterity, we allow a human supervisor to intervene when a trial appears nonrecoverable and cannot proceed to the next stage \cite{goldberg2024augmented}. When this happens, we note the failed suture and allow up to 2 human corrections as seen in the last 2 rows of Fig. \ref{fig:results}. Example human interventions include needle realignment, \kushtimusprime{thread untangling}, thread removal from gripper, or needle replacement after it is dropped.

With respect to human interventions, we found that STITCH 2.0 with 1 intervention averaged 5.55 sutures with a 87.8\% wound gap closure (Table \ref{ablation-table}). STITCH 2.0 with 2 interventions \kushtimusprime{consistently} performed all 6 sutures with a 100\% wound gap closure (Table \ref{ablation-table}).

\subsection{Needle Pose Estimation Ablation}
\begin{figure}[t]
   \centering
   \includegraphics[width=0.9\linewidth]{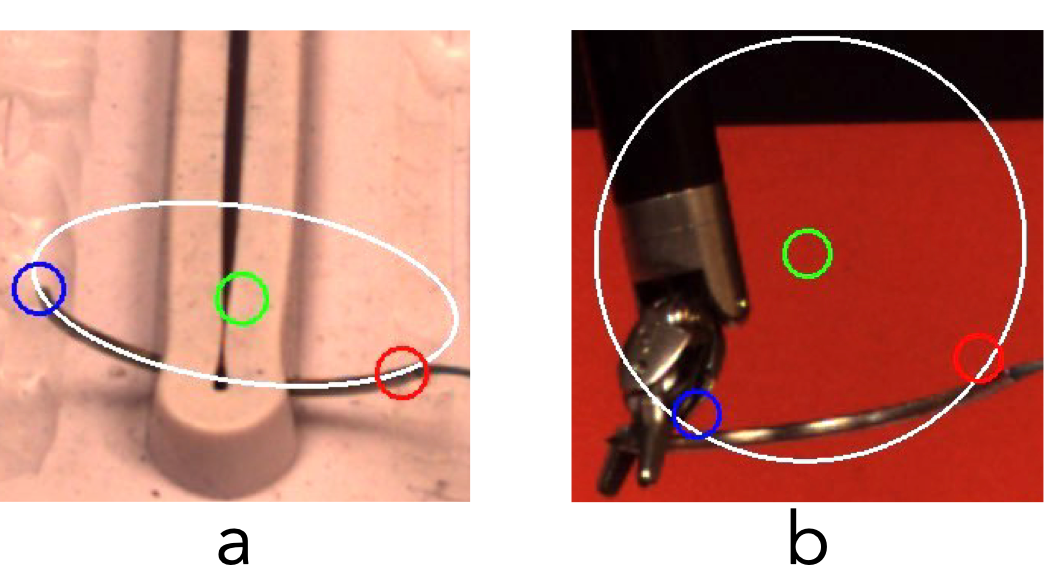}
   \vspace*{-0.1in}
   \caption{\textbf{Needle Pose Estimate Success and Failure.} a. Accurate needle pose estimate: the projected blue and red circles contain the needle endpoints and the projected white circle follows the needle curvature. b. Inaccurate needle pose estimate.}
   \label{fig:needle_pose_eval}
   \vspace*{-0.3in}
\end{figure}

To evaluate needle pose estimation accuracy, we \kushtimusprime{overlaid the needle pose estimate on the image and compared it to the} needle position in the image, with measurements taken prior to each needle grasping or manipulation action during suturing trials. Success and failure examples are shown in Fig \ref{fig:needle_pose_eval}. The complete STITCH 2.0 pipeline achieved 97.56\% accuracy, outperforming the 85.98\% accuracy of the system without EKF \kush{and the 42\% accuracy of STITCH 1.0 (Table \ref{ablation-table}).} 

\kush{The STITCH 2.0 needle pose estimation pipeline performs worse when the needle is oriented away from the camera, where it no longer appears circular in the image, resulting in needle endpoint errors up to 5 mm as shown in Fig \ref{fig:needle_pose_eval}b.}

%% file: Section/limitations.tex
\section{Limitations and Future Work}\label{sec:conclusion}
STITCH 2.0 has three primary limitations. First, the most common failure occurs during needle insertion. Despite accurate needle pose estimation and deep calibration of the robot\cite{hwang2020efficiently}, sub-millimeter gripper inaccuracies sometimes result in the needle tip being at the wrong height for stable insertion. \kush{In future work, we will improve upon the RNN gripper calibration \cite{hwang2020efficiently} to reduce the dVRK kinematic inaccuracies.}

Second, while STITCH 2.0 suture completion time is faster than previous work, it \kushtimusprime{still takes 98.74 seconds per suture when a surgeon could perform a suture in 5-10 seconds.} This is primarily due to needle estimation operating at 1 Hz, limited by the computational demands of stereo reconstruction and pointcloud processing to get stable estimates. Because needle pose estimation relies on 10-15 estimates depending on the state confidence from the Extended Kalman Filter (EKF), needle pose estimation accounts for 63.76\% of total suture time. \kushtimusprime{Future work will explore faster alternatives such as cropping stereo images on the needle to reduce computational load for faster stereo reconstruction and integrating the STITCH 2.0 needle pose estimator with concurrent, realtime monocular keypoint methods \cite{jiang2023markerless}\cite{chiu2022markerless}\cite{Li2024monocular}.}

Third, \kushtimusprime{while STITCH 2.0 addresses challenges in needle pose estimation, thread management, and suture placement relevant for any environment, several important steps are required before STITCH 2.0 can generalize to ex-vivo and clinical settings. Currently, STITCH 2.0 uses an external stereo camera pair rather than an endoscope commonly used in laparoscopic settings. Contemporary stereo endoscopes have a narrow baseline, resulting in noisier pointclouds. Future work will extend STITCH 2.0’s noise reduction methods to better handle pointcloud noise for stereo endoscopes. Additionally, we assume the wound is raised and orthogonal to the camera, which simplifies needle insertion since the needle does not require full rotation through tissue. Additionally, the orthogonal positioning provides clear visibility of insertion and extraction points for suture alignment. Future work will extend the STITCH 2.0 suture alignment and needle placement components to handle more complex wound geometries, including nonlinear and 3D wounds.}